\documentclass{ecai}
\usepackage{times}
\usepackage{graphicx}
\usepackage{latexsym}

\usepackage{amsmath}
\usepackage{multirow}

\usepackage{lipsum}

\begin{document}

\title{Learning Contextualized Sentence Representations for Document-Level Neural Machine Translation}

\author{Pei Zhang   \institute{Soochow University,
China, email: pzhang.nmc@gmail.com, dyxiong@suda.edu.cn} , Xu Zhang \institute{Sogou, China, email: \{zhangxu216526, chenweibj8871, yujian216093, wangyanfeng\}@sogou-inc.com}  , Wei Chen$^2$,    Jian Yu$^2$ , Yanfeng Wang$^2$  \and  Deyi Xiong $^1$  }

\maketitle

\bibliographystyle{ecai}

\begin{abstract}
  Document-level machine translation incorporates inter-sentential dependencies into the translation of a source sentence. In this paper, we propose a new framework to model cross-sentence dependencies by training neural machine translation (NMT)  to predict both the target translation  and  surrounding sentences of a source sentence. By enforcing the NMT model to predict source context, we want the model to learn ``contextualized'' source sentence representations that capture document-level dependencies on the source side. We further propose two different methods to learn and integrate such contextualized sentence embeddings into NMT: a joint training method that jointly trains an NMT model with the source context prediction model and a pre-training \& fine-tuning method that pretrains the source context prediction model on a large-scale monolingual document corpus and then fine-tunes it with the NMT model.   Experiments on Chinese-English  and English-German translation show that both methods can substantially improve the translation quality  over a strong document-level Transformer baseline. 
\end{abstract}

\section{Introduction}
%

Neural machine translation (NMT) has achieved remarkable progress in many languages  due to the availability of the large-scale parallel corpora and powerful learning ability of  neural networks \cite{bahdanau2015neural,sutskever2014sequence,Vaswani2017Attention,gehring2017convolutional}. However, most NMT systems translate a sentence without taking document-level context into account. Due to the neglect of inter-sentential dependencies, even the state-of-the-art NMT models lag far behind human translators on document-level translation \cite{L2018Has}.

Document-level machine translation has been therefore attracting more and more attention in recent years. A variety of approaches have been proposed, which can be roughly divided into two categories: leveraging discourse-level linguistic features \cite{Gong2011Cache,Xiong2013Modeling,Kuang2017Cache} or encoding preceding/succeeding sentences into the model \cite{Zhang2018Improving,Voita2018Context,Hierarchical}. The former may need linguistic resources which are not easily available while the latter  relies on parallel documents. Unfortunately, explicit document boundaries are often removed in parallel texts and it is difficult to automatically recover such boundaries.  

In this paper, we propose a novel approach to document-level neural machine translation. Inspired by the success of contextualized word embeddings in various natural language tasks \cite{Devlin2018BERT,Peters2018Deep,Radford}, we learn contextualized sentence embeddings for document-level NMT. Instead of encoding surrounding sentences into the NMT model, we try to learn a model to predict the previous or next sentence from the current sentence to be translated. This is similar to the skip-thought model \cite{Kiros2015Skip} or the skip-gram model \cite{word2vec} at the word level. By training document-level NMT to predict surrounding source context, we hope the trained encoder to capture document-level dependencies in the sentence embedding of the current source sentence. 

\begin{figure*} [t]
\centering
\includegraphics[height=2.5in,width=5.2in]{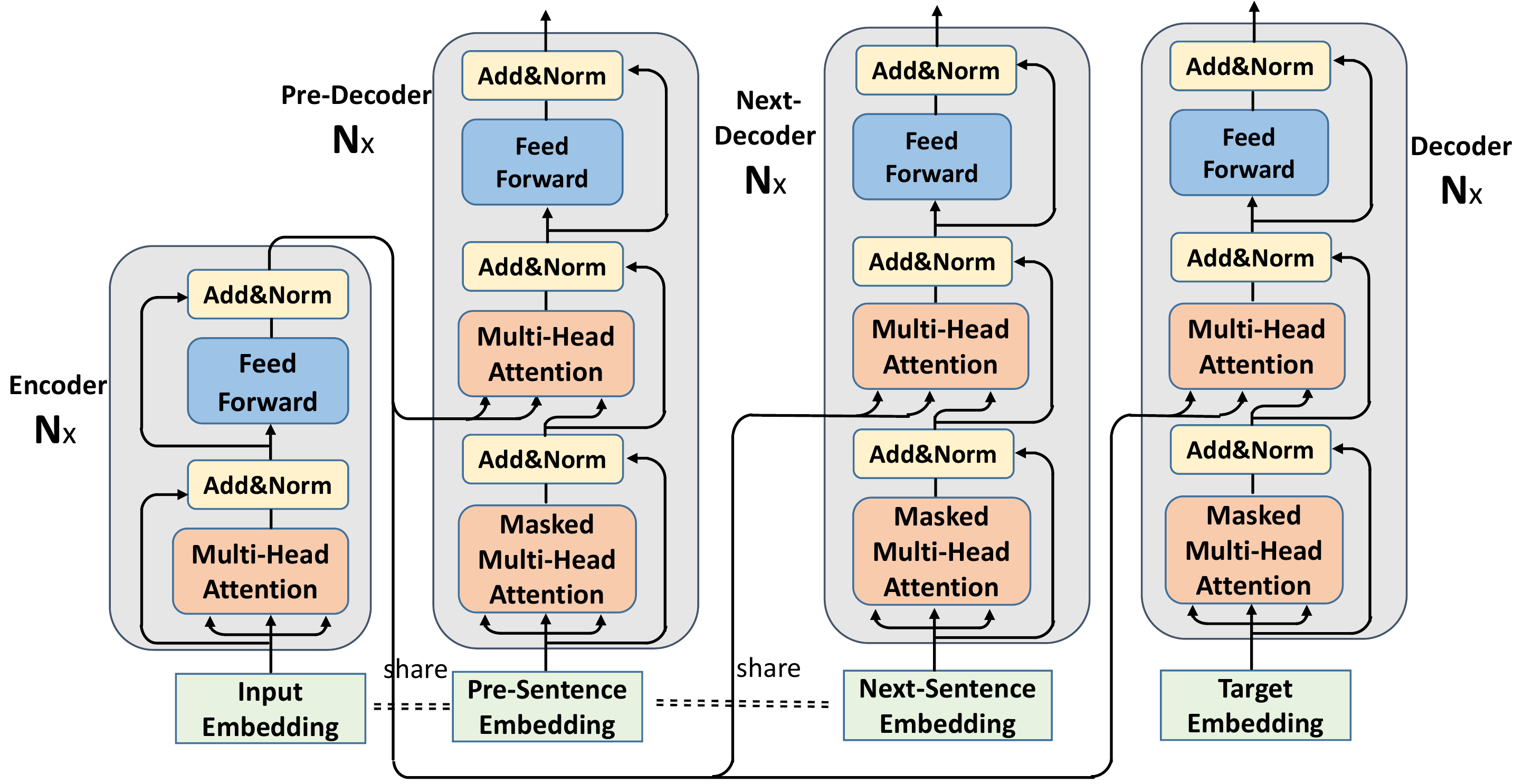}
\caption{Architecture of the joint training model with an encoder learning the  representation of current source sentence  and   three decoders (Pre-Decoder, Next-Decoder and Decoder) for predicting the  previous  source sentence, next source sentence  and target sentences. }
\label{figure of joint-traning Model}
\end{figure*}

\begin{figure*} [t]
\centering
\includegraphics[height=2.5in,width=5.2in]{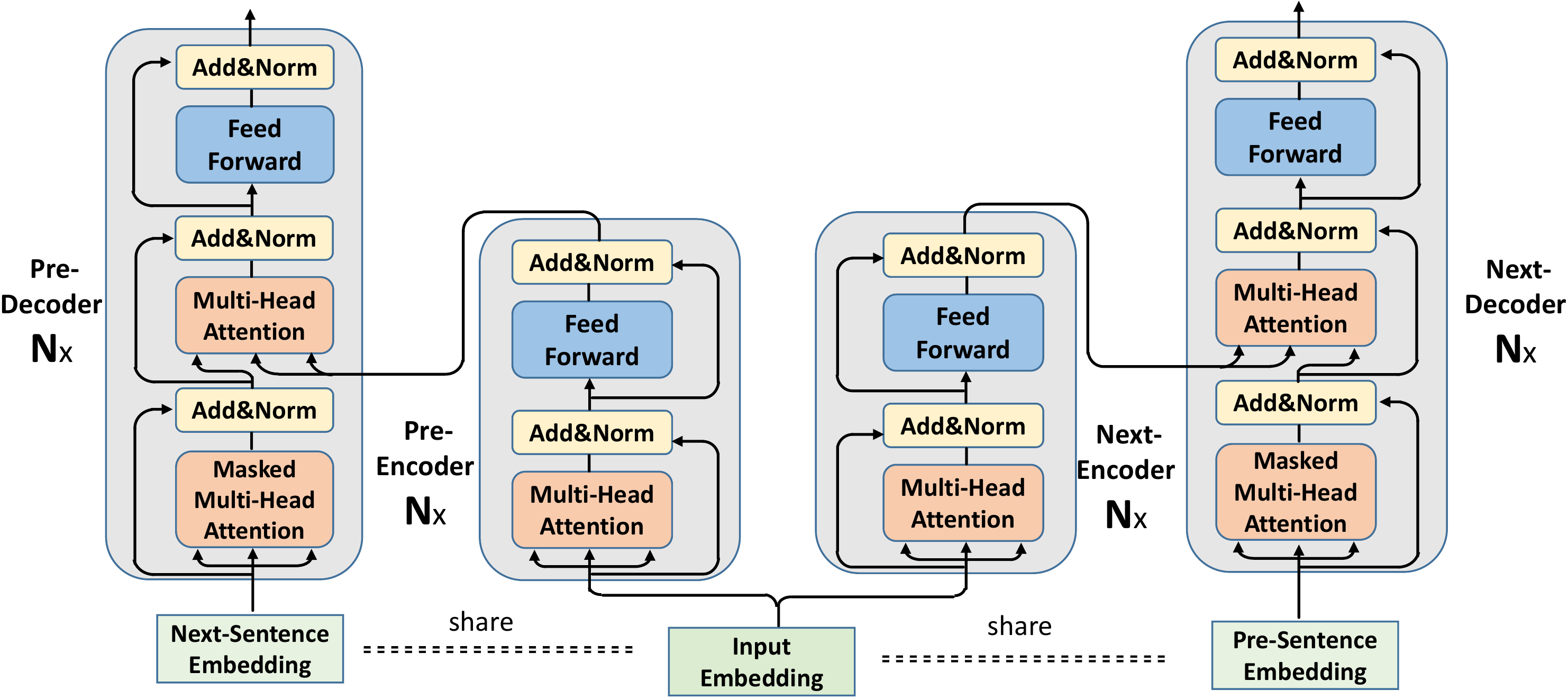}
\caption{Architecture of the pre-training model where two encoder-decoder models are jointly trained to predict the previous and next source sentence from the current source sentence.   }
\label{figure of pre-training Model}
\end{figure*}

\begin{figure*} [t]
\centering
\includegraphics[height=2.5in,width=5.2in]{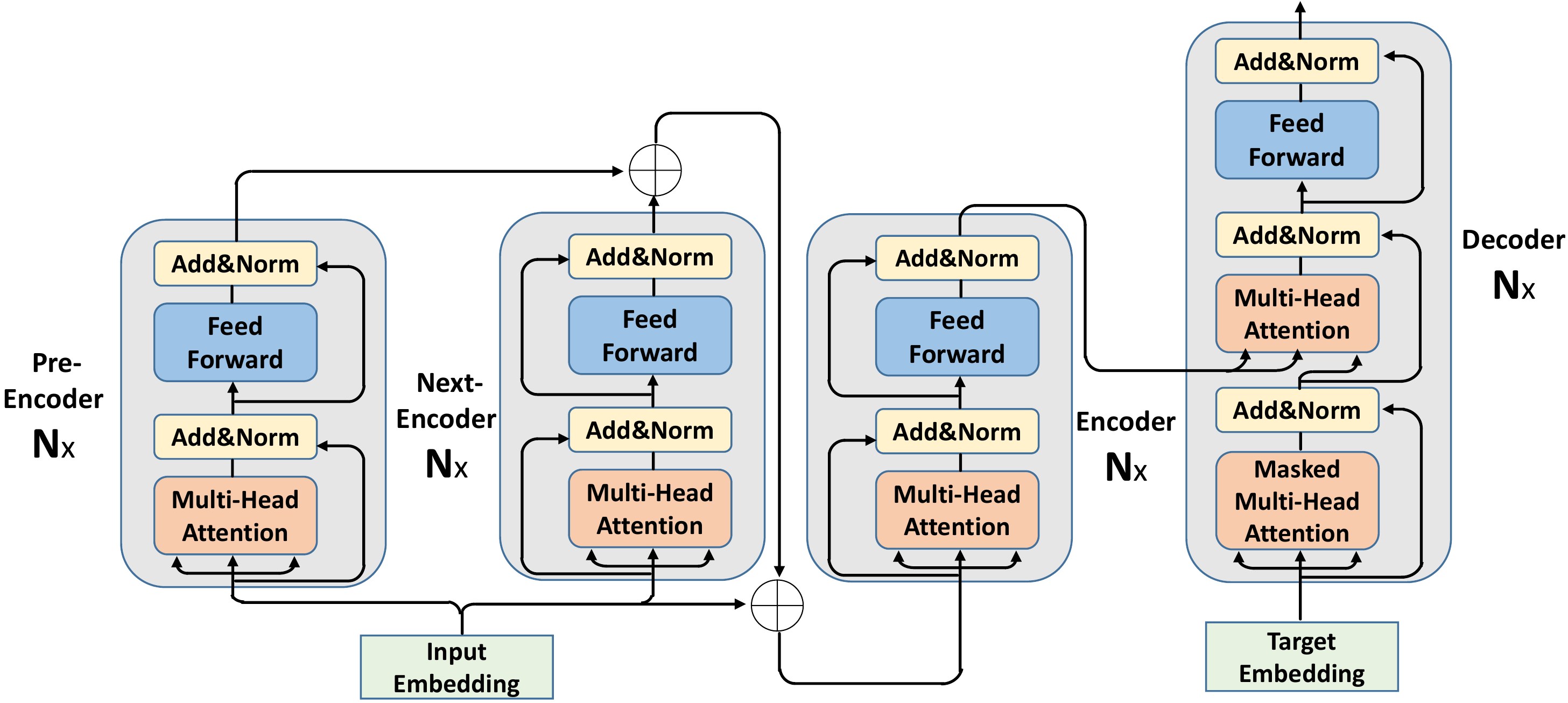}
\caption{Architecture of the fine-tuning model where the sums of the outputs of the pre-trained pre-encoder and next-encoder are input  to the NMT encoder. }
\label{figure of fine-tuning Model}
\end{figure*}

More specifically, we propose two methods to learn and integrate the contextualized sentence embeddings into document-level NMT. First, we explore a joint training method that simultaneously trains an encoder-decoder NMT model with a skip-thought model. As illustrated in Figure \ref{figure of joint-traning Model}, we  use a shared encoder to decode the target sentence, the previous and the next source sentence at the same time. Second, we jointly pretrain two encoder-decoder models to predict the previous and next source sentence respectively from the same current sentence  on a monolingual document-level corpus and fine-tune the pretrained models with the document-level NMT model. In the pre-training \& fine-tuning approach  visualized in Figure \ref{figure of pre-training Model} \& \ref{figure of fine-tuning Model}, the model can use a very  large-scale collection of monolingual corpus with document boundaries, which is easily available. 

Our contributions can be summarized into three aspects: 
\begin{itemize}
\item First, we propose a new framework to document-level NMT by learning contextualized sentence embeddings  on the source side.
\item Second, we further present two approaches to  learning and incorporating contextualized sentence embeddings  into document-level NMT: the  joint training of NMT with a skip-thought model and the combination of pre-training with fine-tuning. 
\item  Third, we validate the effectiveness of the proposed two methods based on the state-of-the-art  NMT architecture Transformer \cite{Vaswani2017Attention} on both Chinese-English and English-German translation.  
\end{itemize}


The paper is organized as follows. Section 2 overviews related work. Section 3 elaborates the two contextualized sentence embedding learning frameworks: the joint training and pre-training with fine-tuning. In section 4, we present the experimental settings and experiment results. In section 5, we conduct in-depth analyses
and discuss our results, followed by our conclusion and future work in section 6.

\section{Related Work}
\label{Related Work}
{\bf Document-level SMT}  Plenty of methods have been proposed for document-level statistical machine translation. 
Gong,  Min,  and  Zhou \cite{Gong2011Cache}
use  a cache-based approach to model document-level machine translation. 
 Meyer  and Popescu-Belis \cite{Meyer2012Using}
explore the discourse connectives to improve the quality of translation. 
 Xiong et al. \cite{Xiong2013Modeling}
propose to learn the topic structure of source document and then map the structure to the target translation. In addition to these approaches  leveraging discourse-level linguistic features for document translation,
Garcia et al. \cite{Retranslationembeddings} 
incorporate new word embedding features into decoder to improve the lexical consistency of translations.

{\bf Document-level NMT} In the context of neural machine translation, previous studies first incorporate contextual information into NMT models built on RNN networks.
Tiede-mann  and  Scherrer \cite{Tiedemann2017Neural}
use  extended source language context to improve the robustness of translation. 
Tu et al. \cite{Tu2017Learning}
propose   a  cache to record the hidden  state of each steps of the  encoder and decoder as contextual  information for word generation.
 Wang  et al. \cite{Wang2017Exploiting}
use a cross-sentence context-aware approach to resolve   ambiguities and inconsistencies of translation. 
Maruf and Haffari \cite{Maruf2017Document} 
propose to use memory networks \cite{Eric2017Key} for document-level NMT. 
Kuang  et  al. \cite{Kuang2017Cache}
propose a cache-based model for document-level NMT, where a static cache is used to store topical words while a dynamic cache \cite{Grave2016Improving} is for words generated in previous translations. 

For document-level NMT based on the Transformer,   
Zhang et al. \cite{Zhang2018Improving} 
propose to explore  previous  sentences  of the current  source sentence as the document information, which is further  exploited by the   encoder and decoder via attention networks.
Xiong  et  al. \cite{Xiong2018Modeling} 
use  multiple   passes of decoding with   Deliberation Network \cite{Deliberation} to improve the translation quality.  When translating the current sentence, translation results of other sentences in the first-pass decoding are used  as the document information. In order to improve translating  anaphoric pronouns, 
Voita et al. \cite{Voita2018Context} 
propose context-aware NMT. Different from these methods, we train document-level NMT to predict surrounding sentences rather than encoding them into NMT  or integrating translations of surrounding sentences into NMT. 
 Miculicich et al. \cite{Hierarchical}
present  a hierarchical attention model to capture  context information and integrate the model into the original NMT.

{\bf Pretrained Language Models} Our work is also related to the recent pretrained language models \cite{Ramachandran2016Unsupervised,Devlin2018BERT,Peters2018Deep,Radford,zhaoetal2018document} that learn contextualized word embeddings.  By learning dynamic context-related embeddings, the pretrained language models significantly improve the representation learning in many downstream natural language processing tasks. In our models, we  also learn ``contextualized'' sentence embeddings by putting a source sentence in its context. We believe that learning the representation of a sentence in its surrounding context is helpful for document-level NMT as the learned representation connection to the surrounding sentences. The way that we learn the ``contextualized'' sentence embeddings is similar to the skip-thought model \cite{Kiros2015Skip} in that we also predict surrounding source sentences from the current source sentence.  But in addition to learning contexualized sentence representations,  we integrate the source context prediction  model into document-level NMT via two methods: the joint training method and the pre-training \& fine-tuning method. 

\section{Learning Contextualized Sentence Embeddings for Document-Level NMT}

\label{Approach}
In this section, we introduce  the   proposed  two methods on the Transformer for document-level NMT.
 \subsection{Joint Training}
 \label{Joint Training}
 When  NMT  translates a source sentence, we want  the sentence representation obtained by the encoder to  contain information that can predict both the target sentence and surrounding source sentences.  For this, we introduce a joint training method over the Transformer to jointly train an NMT model and a Transformer-based skip-thought model, both of which share the same encoder.  Meanwhile, we select the previous and next sentences of the current source sentence as the surrounding sentences to be predicted .
 
 Our joint training  model is shown  in Figure \ref{figure of joint-traning Model}.
 The model is composed of one encoder and three decoders, where the encoder is to encode the source sentence in question and the three decoders are to predict the previous source sentence, next source sentence and target sentence, respectively.  The network stuctures of the encoder and three decoders  are the same as  the Transformer encoder / decoder \cite{Vaswani2017Attention}. In order to train the joint model, we collect   document-level parallel training instances  $(s_{i},s_{i-1},s_{i+1},y_{i})$,  where  $s_{i}$ is the current source sentence. $s_{i-1}$ is the previous sentence of $s_{i}$, $s_{i+1}$ is the next sentence of $s_{i}$ and $y_{i}$ is the target sentence. The task of the joint model is to predict $s_{i-1}$, $s_{i+1}$ and $y_{i}$ at the same time given $s_{i}$. For this, the multi-head attention network between each decoder and the encoder is constructed and automatically learned during  training.  
 
 The  loss function for the joint model is computed as follows:
\begin{gather}
    Loss=Loss_{tgt} +\mu*Loss_{pre} \notag \\
+\lambda*Loss_{next} 
\end{gather}
where $Loss_{tgt}$ is the loss from the  target decoder. $Loss_{pre}$ is the loss of  predicting the previous source sentence and $Loss_{next}$ the next source  sentence. We update model parameters according to the gradient of the joint loss. We use two hyperparameters  $\mu$ and $ \lambda $ for $Loss_{pre}$ and $Loss_{next}$, respectively.  The $s_{i}$, $s_{i-1}$ and $s_{i+1}$ share the same source embeddings.

In order to train  joint model,   we take the following  two steps: 
\begin{itemize}
\item First, we train the joint model to minimize the joint loss function on the collected training set and obtain the best joint model after it converges on the training data. 
\item Second, we remove the  pre-decoder and next-decoder from the joint model and continue to train the reserved NMT encoder and decoder on the parallel training set $\{(s_{i},y_{i})\}$ to optimize  $Loss_{tgt}$.
\end{itemize}

The trained NMT model in this way will be used to examine the effectiveness of the proposed joint training method on the test set.

\begin{table*}
\centering
\caption{\label{BLEU scores in document Chinese-English} BLEU scores of the two methods on Chinese-English translation (trained on the 900K-sentence corpus). ``$\ddagger$'': statistically significantly better than the baseline  $(p<0.01)$.}
\begin{tabular}{|c|l| l l l l l|l|}
\hline \bf Methods & \bf NIST06  & \bf  NIST02 & \bf NIST03 & \bf NIST04 & \bf NIST05  & \bf  NIST08 & \bf AVG    \\ \hline
Baseline & 38.14 & 41.42 & 42.09 & 42.7 & 40.59  & 30.65 & 39.49 \\ \hline
\multicolumn{8}{|c|}{\bf Joint Training} \\ \hline
Pre & 38.63 & 42.65 & 42.77 & 43.05 & 40.84  & 30.81 & 39.96 \\ 
Next & 38.94 & 42.69 & 42.36 & 43.08 & 41.14 & 31.24 & 40.1 \\
Pre+Next & 39.11 $\ddagger$ & 42.72 $\ddagger$  & 43.28 $\ddagger$   & 42.67 & 40.99    & 31.97 $\ddagger$   & 40.33   \\\hline
\multicolumn{8}{|c|}{\bf Pre-training and Fine-tuning} \\ \hline
Pre & 39.39 & 43.04 & 42.62 & 42.83 & 40.78  & 31.9 & 40.23 \\
Next & 39.26 & 42.13 & 42.97 & 42.78 & 41.08 & 32 & 40.2 \\
Pre+Next & 39.11 $\ddagger$ & \bf 43.05 $\ddagger$   & 42.58  & 43.29  & 41.44 $\ddagger$    & 31.73 $\ddagger$  & 40.42  \\
\hline
\multicolumn{8}{|c|}{\bf Pre-training for Joint Training } \\ \hline
Pre+Next  & \bf 39.46 $\ddagger$ & 42.19 $\ddagger$   & \bf 43.39 $\ddagger$  & \bf 43.42    & \bf 41.74 $\ddagger$   & \bf 32.43 $\ddagger$  & \bf 40.63   \\\hline
\end{tabular}

\end{table*}

\begin{table*}
\centering
\caption{\label{BLEU scores in Chinese-English} BLEU scores for the pre-training \& fine-tuning method on Chinese-English translation (trained on the 2.8M-sentence corpus). ``$\ddagger$'': statistically significantly  better than the baseline  $(p<0.01)$. }
\begin{tabular}{|c|l|l l l l l|l|}
\hline \bf Methods  & \bf NIST06 & \bf  NIST02 & \bf NIST03 & \bf NIST04 & \bf NIST05  & \bf  NIST08 & \bf AVG   \\ \hline
Baseline & 43.05 & 43.85 & 44.84 & 46.03 & 43.43  & 36.26 & 42.89 \\ \hline
Pre & 43.48 & 45.2 & 45.76 & 46.34 & 44.45  & \bf 37.16& 43.78\\
Next & 43.15 & 45.01 & 46.3 & 46.3 & 44.91  & 37 & 43.9 \\
Pre+Next  &  \bf 43.62 & \bf 45.78 $\ddagger$& \bf 46.32 $\ddagger$ &  \bf 46.42  $\ddagger$ & \bf 45.23 $\ddagger$ & 37.12 $\ddagger$& \bf 44.17 \\
\hline
\end{tabular}
\end{table*}

\subsection{ Pre-training and Fine-Tuning}
\label{Pre-Traing and Fine-Turing}
The joint  training model requires parallel documents as training data. However, large-scale parallel corpora with document boundaries are not easily available. On the contrary, there are plenty of monolingual documents. Therefore, we further propose a pre-training strategy to train an encoder on monolingual documents. We want the pretrained encoder to capture inter-sentential dependencies by learning to predict surrounding sentences from a current source sentence.

As shown in Figure \ref{figure of pre-training Model}, we jointly pre-train two encoder-decoder models. From a large-scale set of monolingual documents, we can collect a huge amount of triples  $(s_{i-1},s_{i},s_{i+1})$ as training instances for the pre-training model.   The pre-training task trains one encoder-decoder model to predict the previous sentence $s_{i-1}$ from $s_{i}$ and the other $s_{i+1}$ from $s_{i}$.  Source word embeddings are shared by the two encoder-decoder models and they are jointly trained to optimize the following loss: 

\begin{gather}
    Loss=Loss_{pre} + Loss_{next} 
\end{gather}

Once we have the pre-trained model, we can continue to fine tune the pretrained two encoders (pre-encoder and next-encoder) with the Transformer model.  The details are shown in Figure \ref{figure of fine-tuning Model}.  The fine-tuning model contains three encoders and one decoder. The two encoders, namely the pre-encoder and next-encoder that encode the previous and next source sentences during the fine tuning, are from the pretrained model. In order to fine tune the pretrained encoders with the NMT encoder-decoder model,  we use three strategies. First, the source word embeddings of the  pre-encoder, next-encoder and NMT encoder  can be initialized by the word embeddings from the pretrained  model. Second, the input to the NMT encoder is the sum of the outputs of the pre-/next-encoder and word embeddings of the current source sentence, formulated as follows: 
\begin{gather}
   encoder\_input = input\_embedding  \notag \\
            + pre\_encoder\_output \notag \\
            + next\_encoder\_output
\end{gather}
Third, during the fine-tuning stage, the shared source word embeddings and parameters of the pre-encoder and next-encoder continue to be optimized.

\section{Experiments}
We  conducted  experiments  on Chinese-English and   English-German translations  to  evaluate  the effectiveness of the   proposed methods.   
\label{Experiments}

\subsection{Experimental Setting}
For Chinese-English translation, we selected corpora LDC2003E14, LDC2004T07, LDC2005T06, LDC2005T10 and a portion of data from the corpus LDC2004T08 (Hong Kong Hansards/Laws/News) 
 as our bilingual training data, which contain 2.8M sentences. We then  selected from the 2.8M-sentence training data  94K parallel documents with explicit document boundaries,  containing  900K parallel sentences.  Each selected parallel document consists of 11 sentences on average. We used NIST06 dataset as our development set and NIST02, NIST03, NIST04, NIST05, NIST08 as our test sets. The development and test  datasets contain 588 documents and 5,833 sentences  in total. Each document has 10 sentences averagely. We also collected  a large-scale monolingual (Chinese) document corpus from CWMT\footnote{http://nlp.nju.edu.cn/cwmt-wmt.} and Sogou Labs\footnote{ https://www.sogou.com/labs/resource/list\_news.php.}, with  25M sentences and 700K documents. On avegrage, each documents contains 35 sentences.
 
For English-German translation, we used the
WMT19 bilingual document-level training  data\footnote{https://s3-eu-west-1.amazonaws.com/tilde-model/rapid2019.de-en.zip.}, which contains  39k documents with 855K sentence pairs as training set.  We collected a large-scale English monolingual document corpus\footnote{http://data.statmt.org/news-crawl/en-doc/news-docs.2015.en.filtered.gz.} from WMT19 with 10M sentences and  410k documents. We used the newstest2019 development set as our development set  and newstest2017, newstest2018 as test sets, which contain 123 documents with 2,998 sentence pairs and 252 documents with 6,002 sentence pairs respectively.

We used the  byte pair encoding  \cite{Sennrich2015Neural} to decompose words  into small sub-word units for both languages. We used the case-insensitive 4-gram NIST BLEU score as our evaluation metric \cite{papineni2002bleu} and the script ``mteval-v11b.pl'' to compute BLEU scores. All the out-of-vocabulary words were replaced with a  token ``UNK''.

As mentioned before, we used the Transformer to construct our models and   implemented our models on an  open source toolkit THUMT \cite{Zhang2017THUMT}. We set hidden size to 512 and filter-size to 2,048. The number of encoder  and decoder layers was 6 and the number of attention heads was 8. We used Adam \cite{Kingma2014Adam} for optimization. The learning rate was set to  1.0 and  the number of warm-steps was 4000. We set batch size as 4,096 words for iterations. We used four TITAN 

\begin{table}
\centering
\caption{\label{BLEU scores Two step} BLEU scores (average results on the 5 test sets) for the two-step training  strategy for the  ``joint training'' and  ``Pre-Training + Joint Training'' methods abbreviated into ``PT + JT'' for limited  space on the 900K-sentence corpus. }
\begin{tabular}{|c|c|c|}
\hline \bf Methods   & \bf  Joint Training & \bf PT +  JT  \\ \hline
Pre+Next (step 1)     & 40.12 & 40.35  \\ \hline
Pre+Next (step 2)     & 40.33 & 40.63 \\
\hline
\end{tabular}
\end{table}

\begin{table}
\centering
\caption{\label{BLEU scores in additional context Representation } Comparison between the joint training method and the explicit contextual integration method on Chinese-English translation.}
\begin{tabular}{|c|l| l l l l l|l|}
\hline \bf Methods &  \bf AVG    \\ \hline
Baseline  & 39.49 \\ \hline
Joint Training & 40.33 \\ \hline
Explicit Contextual Integration  & 40.59  \\ \hline
\end{tabular}
\end{table}

\noindent Xp GPUs for training and two TITAN Xp GPUs for decoding.  Additionally, during decoding, we used the beam search algorithm and set the beam size to 4.  We used bootstrap resampling method \cite{significance} to conduct statistical significance test on results.

We compared our models against the Transformer \cite{Vaswani2017Attention} as our baseline and some previous document-level NMT models 
\cite{Zhang2018Improving,Tiedemann2017Neural}.

\subsection{The Effect of  the Joint Training}
The experimental results of  the joint training method are shown  in Table \ref{BLEU scores in document Chinese-English}. ``Pre'' / ``Next'' indicate that we use only the pre-decoder / next-decoder  in the joint training model illustrated in Figure \ref{figure of joint-traning Model}.  We set $\mu=0.5$ and $\lambda=0.5$ for the ``Pre'' and ``Next'' methods  according to the experiment results on the development set. The BLEU scores of these two methods are close to each other,  indicating that   the previous and next sentences have almost the same influence on the translation of the current sentence.  When we use ``Pre+Next'' method to predict  the  previous and next sentence at the same time  ( $\mu=0.5, \lambda=0.3 $ set according to results on the development set), the performance  is  better than  ``Pre'' and ``Next'' alone achieving an improvement of +0.84 BLEU points  over the baseline. 

\subsection{The Effect of the Pre-training \&  Fine-Tuning}
We trained the  pre-training model on the 25M-sentence monolingual document corpus and then conducted the fine-tuning as shown in Figure \ref{figure of fine-tuning Model} on the two different parallel datasets: the 900K-sentence parallel corpus and the 2.8M-sentence parallel corpus. Sentences from the same document in the former corpus exhibit strong contextual relevance to each other. However, in the latter corpus,  not all documents  have clear document boundaries.  When we train the  pre-training model  on the monolingual document corpus, we did not shuffle sentences in documents and kept the original order of sentences in  each documents. The results are shown in Table ~\ref{BLEU scores in document Chinese-English} and Table ~\ref{BLEU scores in Chinese-English}. Similar to the joint training, we can train the pre-training model with a single encoder, either the pre-encoder or the next-encoder, to obtain the results for ``Pre'' or ``Next''. Of course, we can train the two encoders together to have the results for ``Pre+Next''. As shown in Table \ref{BLEU scores in document Chinese-English} \& \ref{BLEU scores in Chinese-English}, the ``Pre'' and ``Next'' in the pre-training \& fine-tuning model obtain comparable improvements over the two baselines. The improvements over the Transformer baseline without using any contextual information are +0.93 and +1.28 BLEU points on the two corpora, respectively. 

\begin{table}
\centering
\caption{\label{BLEU scores Previous} Comparison to other document-level NMT methods on Chinese-English translation. }
\begin{tabular}{|c|c|c|}
\hline \bf Methods   & \bf 900k & \bf 2.8M   \\ \hline
Baseline & 39.49 & 42.89\\  \hline
(Tiedemann and Scherrer \cite{Tiedemann2017Neural}) & 38.83 & - \\ \hline
(Zhang et al. \cite{Zhang2018Improving}) & 39.91 & 43.52 \\ \hline
Our work & 40.63 & 44.17 \\ \hline
\end{tabular}
\end{table}

\subsection{The Effect of the Pre-training for the Joint Training}
Both the  joint training and pre-training are able to improve the performance in our experiments.  We further conducted experiments to test the combination of them.   Particularly, in the combination of two methods, we used  the source embeddings learned from the  pre-training model to initialize the  source embeddings of the joint training model.  It can be seen from Table \ref{BLEU scores in document Chinese-English} that  such a combination of the two methods achieves the  highest BLEU scores with +1.14 BLEU points higher than the Transformer baseline and  +0.3 BLEU points higher  than the  single  joint training model.  

\subsection{The Effect of the Two-Step training for the Joint Training}
As mentioned in Section \ref{Joint Training}, we take two steps to train the joint training model. Here we carried out experiments to examine this two-step training method. Results are shown in Table \ref{BLEU scores Two step}, from which we can clearly see that the two-step training method is able to improve both the single joint training model and the combination of the joint training with the pre-training model.

\subsection{Comparison to the Explicit Integration of Preceding/Succeeding Source Sentences into NMT }
In our joint training method, during the testing phase, we do not explicitly use any contextual information for the current source sentence as the jointly trained NMT model implicitly learns the potential context information by learning to predict the preceding / succeeding source sentence during the training phase. In order to study how much potential context information can be captured by this implicit method, we conducted a comparison experiment. We obtain the representations of the preceding / succeeding source sentences via the pre-encoder and next-encoder  and integrate them into the encoder of the current sentence by using self-attention that treats the encoder outputs of the current sentence as $q$ and the encoder outputs of the previous / next sentence as $k$ and $v$. We refer to this method as ``Explicit Contextual Integration". The results are shown in Table ~\ref{BLEU scores in additional context Representation }. The explicit contextual integration method is better than our joint training method by only 0.26 BLEU points. Comparing this with the improvement of 0.84 BLEU points achieved by our joint training method over the baseline, we find that the joint training method is able to learn sufficient contextual information for translation in an implicit way.

\subsection{Comparison with Other Document-Level NMT Methods}
We further conducted experiments to compare our methods  with the following previous document-level methods on Chinese-English translation: 

\begin{table}
\centering
\caption{\label{BLEU scores EN-DE} BLEU scores on English-German translation. }
\begin{tabular}{|c|c|c|}
\hline \bf Methods   & \bf BLEU   \\ \hline
Baseline & 17.08  \\  \hline
Joint Training & 17.89  \\ \hline
Pre-training \& Fine-tuning & 18 \\ \hline
\end{tabular}
\end{table}

\begin{table}
\centering
\caption{\label{BLEU scores Two Encoders Not Sharing Parameters } BLEU scores (average results on the 5 test sets) of the two encoders vs. the single shared encoder for the pre-training model on the 900K-sentence corpus.  }
\begin{tabular}{|c|c|}
\hline \bf Methods   & \bf  BLEU    \\ \hline
Baseline & 39.49  \\ \hline
Single Shared  Encoder & 39.95  \\ \hline
Two Encoders & 40.42  \\ \hline
\end{tabular}
\end{table}

\begin{itemize}
\item  (Tiedemann and Scherrer \cite{Tiedemann2017Neural}): concatenating  previous sentence with the current sentence and inputting them into the encoder.
\item  (Zhang et al. \cite{Zhang2018Improving}): using  previous  sentences  of the current  source sentence as  document information, which is further  exploited by the   encoder and decoder via attention networks. 
\end{itemize}

Table \ref{BLEU scores Previous} shows the comparison results in terms of BLEU scores. It is clear that our method outperforms these two methods. 
 Although the concatenation method \cite{Tiedemann2017Neural} can improve document-level NMT over the  RNNSearch model \cite{o2017The},   it fails to improve the Transformer model. Comparied with the method by 
 Zhang  et  al. \cite{Zhang2018Improving} 
 that is also based on Transformer,  our work outperforms their model  by 0.72 and 0.65 BLEU points on the two datasets.
 
\subsection{Results on English-German Translation}
The results are shown in Table \ref{BLEU scores EN-DE}, which shows that our two methods also outperforms the baseline  by 0.81 and 0.92 BLEU points on English-German translation, respectively.

\section{Analysis}
\label{Analysis}
Experiments on the  two methods show that using the context of previous and next sentence for document translation  can improve NMT performance. In this section, we provide further analyses and discussions on the two methods.

\subsection{Analysis on the   Two Encoders for the Pre-training Model}
As we have described in Section \ref{Pre-Traing and Fine-Turing}, we use two separate encoders in the pre-training model to encode the current source sentence for predicting the previous and next sentence.
Only the word embeddings are shared in these two encoders.  An alternative to these two encoders is to use a single encoder that is shared by the pre-decoder and next-decoder. 

We conducted experiments to compare the two encoders vs. the single shared encoder for the pre-training model. Results are shown in Table  \ref{BLEU scores Two Encoders Not Sharing Parameters }. Obviously, the two-encoder network is better than the single shared encoder. This indicates  that the two separate encoders are better at capturing the discourse dependencies between the previous and the current sentence and between the next and the current sentence than the single encoder. We conjecture that this is because the dependencies on the previous sentence are different from those on the next sentence. 

\begin{table}
\centering
\caption{\label{BLEU scores Input Embedding} BLEU scores for the ablation study of the fine-tuning model on the two corpora.  ``Pre+Next+Input Embedding'' is the method that inputs the sum of  the  outputs from the  pre- and next-encoder  and input embeddings into the fine-tuning model. ``Input Embedding'' only uses input embeddings of the pre-training model for the input to the  fine-tuning  model. }
\begin{tabular}{|c|c|c|}
\hline \bf Methods   & \bf  900k & \bf 2.8M  \\ \hline
Baseline & 39.49 & 42.89  \\  \hline
Input Embedding & 39.83 & 43.55 \\ \hline
Pre+Next+Input Embedding & 40.33 & 44.17 \\ \hline
\end{tabular}
\end{table}

\begin{table}
\centering
\caption{\label{BLEU Additional parameters} The number of parameters used in the baseline and our method and performance comparison.  }
\begin{tabular}{|c|c|c|}
\hline \bf Methods   & \bf Parameters & \bf BLEU   \\ \hline
Baseline & 90.2M & 39.49 \\ \hline
Joint Training & 90.2M & 40.33 \\  \hline
Pre-training \& Fine-tuning & 128M & 40.42 \\ \hline
No Pre-training & 128M & 38.8 \\ \hline
\end{tabular}
\end{table}

\subsection{ Ablation Study for the Fine-Tuning Model}
As we mentioned in Section \ref{Pre-Traing and Fine-Turing},  we use the sum of  input word embeddings, the output of the   pre-encoder and next-encoder loaded from the  pre-training model  as the input to the NMT encoder. Here we empirically investigate the impact of  these three parts on the fine-tuning model via ablation study. From the results shown in Table \ref{BLEU scores Input Embedding}, we can find that if we use only the pretrained word embeddings as the input word embeddings to the NMT encoder of the fine-tuning model, we can obtain improvements of +0.34 and +0.66 BLEU points over baseline on the two corpora. When we add the outputs from both the pre-encoder and next-encoder, we obtain further improvements of +0.5 and +0.62 BLEU points over the input word embeddings. This clearly suggests that the pretrained two encoders learn additional   contextual information. 

\subsection{Analysis on the  Additional Parameters}
Table \ref{BLEU Additional parameters} shows the number of parameters used by the baseline and our method. The joint training method uses the same number of parameters as the baseline, achieving a higher BLEU score. The number of parameters in the pre-training \& fine-tuning method is about 30\% larger than that of the baseline due to the additional two encoders for the preceding and succeeding source sentence. If we do not pre-train the model, the performance significantly drops, even worse than the baseline. This suggests that the improvement of the pre-training \& fine-tuning method over the baseline is achieved not because of using additional parameters but the learned contextual knowledge in the pre-training on the monolingual document data. 

\begin{table*}
\centering
\caption{\label{Translation} Translation examples of the baseline Transformer and our best model. }
\begin{tabular}{c|l} \hline
\hline  SRC   &   dang ran   ,     mu qian  fei zhou   di qu   de   \emph{ \textbf{ wen ding   ju mian}}   yi ran   bi jiao   \emph{ \textbf{cui ruo}}   .     \\ \hline
REF & of course the \emph{ \textbf{stability}} in africa at present is still  \emph{ \textbf{fragile}} . 
  \\ \hline
Transformer & of course , the current \emph{ \textbf{situation}} in africa is still relatively \emph{ \textbf{weak}} . 
  \\ \hline
Our best model & of course , the current \emph{ \textbf{stability}} in the african region is rather \emph{ \textbf{fragile}} .
  \\ \hline \hline
  
 \multirow{3}*{SRC} & (1) zai xiang gang   jing wu chu   du pin   ke   zhi chu   ,   zhe   shi   xiang gang jing fang   de  shou ci   \emph{ \textbf{fa xian}} .  \\
        ~ &  (2) xiang gang   jing fang   chen   ,   zai   guo qu   de   yi   nian   zhong   ,   gong   \emph{ \textbf{fa xian}}  
         shu   zong  an jian   . \\ \hline
 \multirow{4}*{REF} & (1) the dangerous drug division of hong kong police department points out that this is the  \\    
    ~ &  first \emph{ \textbf{discovery}} by the police .    \\
        ~ &  (2) according to hong kong police, they have \emph{ \textbf{discovered}} several cases last year .   \\ \hline
 \multirow{2}*{Transformer} & (1) hkp  pointed out that this was the first time that the hong kong police had \emph{ \textbf{found}} . \\
        ~ &  (2) the hong kong police said that in the past year , a number of cases  were \emph{ \textbf{detected}} . \\ \hline
\multirow{2}*{Our best model} & (1) hkp pointed out that this was the first time that the hong kong police ( hkp )   had \emph{ \textbf{detected}} it . \\ 
        ~ &  (2) according to the hong kong police , in the past year , a number of drug cases were \emph{ \textbf{detected}} .
 \\ \hline \hline
\end{tabular}
\end{table*}

\subsection{ Analysis  on  Translations  }
We take a deep look into the translations generated by the baseline Transformer and our best model. We find that our methods can improve translation quality by disambiguating word senses with document context (as shown in the first example in Table \ref{Translation}) or by making the document-level translations more consistent (as illustrated in the second example in Table \ref{Translation})  and so on. In the first example, ``cui ruo'' is ambiguous with two senses of ``weak'' or ``fragile'', which can be translated correctly by our model since it has learned the exactly  contextualized sentence embeddings. In  the second example, our model translates ``fa xian'' in both sentences into the same translation ``detected", because  the meaning of the ``an jian'' (case) mentioned by the second sentence can be predicted from the meanings of ``du pin'' (drug) and ``jing fang'' (police) in the first sentence.  

We  made a further statistical analysis  on  these  two kinds of translation improvements. We randomly selected  five documents with 48 sentences in total from the  test datasets for translation. Among  these sentences, our model can successfully deal with translations of ambiguous words in 32 sentences.  For the document-level consistent translation  problem, we found that 4 sentences among 7 sentences with such consistent phenomena were correctly and consistently translated by our model.

\section{Conclusion and Future Work}
\label{Conclusion and Future Work}
In this paper, we propose a new framework to document-level NMT by learning a contextualized representation for a source sentence to be translated. We have presented two methods for learning and integrating such representations into NMT. The joint training method learns contextualized sentence embeddings simultaneously with the prediction of target translations. The pre-training \& fine-tuning method learns the contextualized representations on a  large-scale monolingual document corpus and then fine-tunes them  with NMT. Experiments and analyses validate the effectiveness of the proposed two methods.

The proposed two methods can be extended in several ways. First, we would like to train the pre-training model with more monolingual documents so as to learn better contextualized sentence representations. Second, we also want to adapt the current framework from the source side to the target side in the future.

\bibliography{ecai}
\end{document}